\documentclass[preprint,12pt,authoryear,longtitle]{elsarticle}

\usepackage{amssymb}
\usepackage{amsmath}

\usepackage{hyperref}

\usepackage{graphicx}
\usepackage{booktabs}

\usepackage[font=footnotesize]{subcaption}

\usepackage[separate-uncertainty=true, per-mode=symbol]{siunitx}
\usepackage{tabularx}
\usepackage{mathtools}
\usepackage{dsfont}
\usepackage{pifont}

\usepackage{xcolor,colortbl}
\definecolor{Gray}{gray}{0.85}
\newcolumntype{a}{>{\columncolor{Gray}}c}

\journal{Neural Networks}

\begin{document}

\begin{frontmatter}



\title{GMM-COMET: Continual Source-Free Universal Domain Adaptation via a Mean Teacher and Gaussian Mixture Model-Based Pseudo-Labeling} 


\author[1]{Pascal Schlachter\corref{cor1}\fnref{fn1}}
\ead{pascal.schlachter@iss.uni-stuttgart.de}
\author[1]{Bin Yang\fnref{fn2}} 
\ead{bin.yang@iss.uni-stuttgart.de}
\affiliation[1]{organization={Institute of Signal Processing and System Theory, University of Stuttgart},
            addressline={Pfaffenwaldring 47}, 
            city={Stuttgart},
            postcode={70569}, 
            country={Germany}}
        
\cortext[cor1]{Corresponding author}
\fntext[fn1]{CRediT roles: Conceptualization, Methodology, Software, Investigation, Writing - Original Draft, Visualization}
\fntext[fn2]{CRediT roles: Supervision, Writing - Review \& Editing}

\begin{abstract}
Unsupervised domain adaptation tackles the problem that domain shifts between training and test data impair the performance of neural networks in many real-world applications. Thereby, in realistic scenarios, the source data may no longer be available during adaptation, and the label space of the target domain may differ from the source label space. This setting, known as source-free universal domain adaptation (SF-UniDA), has recently gained attention, but all existing approaches only assume a single domain shift from source to target. In this work, we present the first study on continual SF-UniDA, where the model must adapt sequentially to a stream of multiple different unlabeled target domains. Building upon our previous methods for online SF-UniDA, we combine their key ideas by integrating Gaussian mixture model-based pseudo-labeling within a mean teacher framework for improved stability over long adaptation sequences. Additionally, we introduce consistency losses for further robustness. The resulting method GMM-COMET provides a strong first baseline for continual SF-UniDA and is the only approach in our experiments to consistently improve upon the source-only model across all evaluated scenarios. Our code is available at \url{https://github.com/pascalschlachter/GMM-COMET}.
\end{abstract}

\begin{graphicalabstract}
\includegraphics[width=\linewidth]{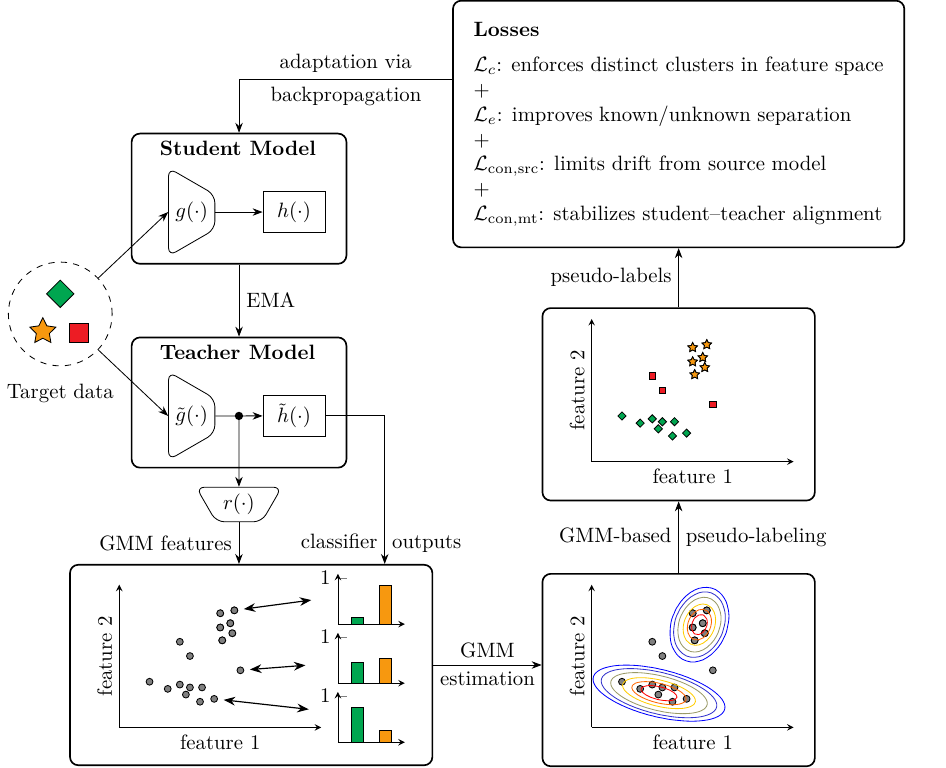}
\end{graphicalabstract}

\begin{highlights}
	\item We are the first to study the challenging setting of continual source-free universal domain adaptation (SF-UniDA), addressing both evolving domains and a shift in the label space while being strictly source-free.
	\item Building on our previous contributions, we integrate the Gaussian mixture model (GMM)-based pseudo-labeling \citep{GMM} with the mean teacher framework of COMET \citep{COMET}, forming a unified approach that combines the strengths of both methods to effectively handle continual adaptation. We call the resulting framework GMM-COMET.
	\item Beyond merely combining the basic ideas of our two previous approaches towards online SF-UniDA, we incorporate new consistency losses into GMM-COMET to promote stable performance and mitigate drift over time, strengthening the method against the challenges of long-term adaptation in the continual setting.
	\item We provide the first comprehensive evaluation of continual source-free universal domain adaptation and establish a strong initial benchmark for future research. By achieving the most consistent performance across the diverse domain and category shift scenarios, our proposed framework GMM-COMET outperforms all existing methods.
\end{highlights}

\begin{keyword}
Universal Domain Adaptation \sep Continual Test-Time Adaptation \sep Source-Free Domain Adaptation \sep Contrastive Learning \sep Mean Teacher \sep Gaussian Mixture Model



\end{keyword}

\end{frontmatter}



\section{Introduction}
Despite their recent success, deep neural networks still struggle when the training and test data $(x, y) \in \mathcal{X}\times\mathcal{Y}$ are not drawn i.i.d. from the same distribution $\mathcal{P}$ on $\mathcal{X}\times\mathcal{Y}$ \citep{iid_assumption}. Here $\mathcal{X}$ represents the input space, and $\mathcal{Y}$ the label space. This issue arises due to a domain (distribution) shift between the training and test data, which is common in real-world applications. In such cases, while $\mathcal{X}$ and $\mathcal{Y}$ remain the same, the distribution $\mathcal{P}_\mathrm{t}$ of the target domain differs from the distribution $\mathcal{P}_\mathrm{s}$ of the source domain. To address this domain shift, unsupervised domain adaptation (UDA) is commonly applied. UDA adapts a pre-trained model from the labeled source domain to the unlabeled target domain, aiming to minimize the performance drop caused by the distribution mismatch. To achieve this, traditional UDA methods leverage both the source and target data during adaptation. However, in many real-world scenarios, accessing the source data after model deployment is not feasible due to privacy concerns, proprietary restrictions, or storage limitations. To overcome these challenges, source-free UDA, also known as test-time adaptation (TTA), has emerged, where adaptation is performed using only the unlabeled target data. This approach not only preserves data privacy but also enhances computational efficiency, as it eliminates the need to reprocess the source data during adaptation \citep{tent}.

Nevertheless, most source-free UDA approaches still operate under the assumption that the label space remains identical across both the source and target domains, which is referred to as the closed-set setting. However, in many real-world applications, the target label space differs from that of the source, leading to a category shift in addition to the domain shift. This shift occurs when new classes appear in the target domain that were absent in the source domain, or when certain source domain classes are missing in the target domain. Specifically, three types of category shift scenarios are possible: partial-set DA (PDA), open-set DA (ODA) and open-partial-set DA (OPDA). Since prior knowledge of the target domain's class composition is typically not available, source-free universal domain adaptation (SF-UniDA) was introduced which aims to universally perform well across all category shift scenarios \citep{glc}. Accordingly, it needs to correctly classify samples belonging to the known classes present in both the source and target domains, and to reject samples from new, unseen classes by labeling them as ``unknown''.

In this sense, SF-UniDA shares similarities with few-shot learning \citep{few_shot} and class-incremental continual learning \citep{continual_learning_survey} as both also aim to adapt a pre-trained model to previously unseen classes. However, unlike these approaches, SF-UniDA operates without any labeled target data, focusing on only rejecting unseen classes as ''unknown`` instead of finding labels for them. Additionally, SF-UniDA addresses both domain and category shifts simultaneously, whereas few-shot and class-incremental continual learning only aim to adapt to new classes without considering a domain shift.

Another closely related direction is zero-shot learning, which especially in the form of vision–language models (VLMs) such as CLIP enables recognition of previously unseen classes (e.g. \citep{zero_shot_1, zero_shot_2}). These models do not require labeled target data and often generalize well across domains due to large-scale pre-training. However, this capability comes at the cost of substantial model size and computational requirements, making VLMs unsuitable for many embedded or resource-constrained applications. For this reason, SF-UniDA is mainly relevant for compact models, which is why we focus solely on non-foundation models in our work.

In addition to handling category shifts, many real-world applications require models to continuously adapt to incoming data streams. This scenario, known as online TTA, demands adaptation to be performed simultaneously with prediction to enable real-time inference. In contrast, traditional offline TTA assumes a finite target dataset and performs the adaptation entirely before inference. While online TTA only addresses a single domain shift between the source and target domain, continual TTA \citep{CoTTA} goes one step further by tackling a sequence of multiple consecutive domain shifts within the target domain. This scenario is more realistic but also more challenging, as adapting to long test sequences often leads to performance degradation over time due to the accumulation of prediction errors and catastrophic forgetting \citep{CoTTA}. Moreover, domain shifts occur unpredictably, with no prior information about when they will happen.

While we recently introduced the first methods for online SF-UniDA \citep{COMET, GMM}, the continual SF-UniDA setting remains unexplored. In this paper, we make a substantial step forward by unifying and extending the core ideas of both prior approaches into a novel method we call GMM-COMET to address the more demanding continual SF-UniDA scenario. Beyond methodological advancements, we conduct comprehensive experiments and establish the first benchmark for this setting. Notably, our method GMM-COMET is the only one that consistently outperforms the source-only baseline across all three category shift types and three standard domain adaptation datasets. In this way, it achieves the best overall performance among all evaluated methods. 

\section{Related Work}
\subsection{Continual Test-Time Adaptation}
Continual TTA was first studied by \citep{CoTTA}. They propose using a mean teacher for pseudo-labeling preventing error accumulation and catastrophic forgetting during the continual adaptation. To enhance pseudo-label quality, they average teacher predictions over multiple augmentations for large domain shifts. The student model is then updated using a cross-entropy (CE) loss. Additionally, to preserve source knowledge and reduce model drift, they stochastically reset a small number of model weights to their original source values.

Subsequent methods have largely retained this mean teacher framework, modifying or extending it. For instance, \citep{CTTA_DSS} proposes Dynamic Sample Selection (DSS) to split target samples into high- and low-confidence groups. High-confidence samples are then used for ``positive learning'' implemented as a standard CE loss between student and teacher predictions. Low-confidence samples, in contrast, are used for ``negative learning'', where a cross-entropy loss with complementary labels pushes the student prediction away from classes with low confidence predictions. \citep{rmt} replaces the standard loss with a symmetric cross-entropy (SCE), which offers more robust gradient behavior. They further incorporate a contrastive loss and employ source replay to combat forgetting. Similarly, \citep{ccotta} builds on the mean teacher with SCE loss but introduces feature-space shift control, using concept activation vectors derived from class prototypes to prevent class overlap and limit the overall distributional shift.

\subsection{Universal Domain Adaptation}
Most existing methods addressing a combined domain and category shift are not source-free, as they either rely directly on source data for adaptation \citep{cao2018partial, cao2018partial2, zhang2018importance, open_set_domain_adaptation, saito2018open, psdc, liu2019separate, busto2018open, baktashmotlagh2018learning, bucci2020effectiveness, you2019universal, fu2020learning, saito2021ovanet, saito2020universal, Li_2021_CVPR, chen2022geometric, lu2024mlnet} or require a dedicated, e.g. open-set, source training and/or architecture already preparing the source model for an anticipated category shift \citep{universal_sf_da, kundu2020towards, umad, coca}. While some of these latter methods avoid reprocessing source data during adaptation, they still rely on additional source knowledge beyond a standard pre-trained source model like source data, a dedicated (e.g. open-set) source training or/and a specific source model architecture. This significantly limits their practical applicability. Therefore, adhering to the strict definition of source-free adaptation outlined in \citep{glc}, we do not classify these methods as truly source-free. Furthermore, some of the remaining source-free methods are not universal, as they either target a single specific category shift \citep{owttt, feng2021open} or require prior knowledge to adapt the approach to a particular shift \citep{shot}.

Only \citep{glc, glc++, lead, COMET, GMM} can be considered both truly source-free and universal. \citep{glc, glc++} propose a pseudo-labeling approach called Global and Local Clustering (GLC). The generated pseudo-labels are utilized to optimize a combination of losses: a cross-entropy loss, a kNN-based loss, and, in the case of GLC++ \citep{glc++}, an additional contrastive loss. The rejection of unknown classes during inference is realized by an entropy threshold. \citep{lead} builds on this work but replaces the pseudo-labeling by a novel approach called Learning Decomposition (LEAD). It separates features into source-known and -unknown components, enabling instance-level decision boundaries to distinguish between known and unknown classes during pseudo-labeling. Moreover, they add a confidence weighting for the cross-entropy loss and use a feature decomposition regularizer instead of the contrastive loss. However, we recently demonstrated that these methods, designed for offline SF-UniDA, perform poorly in the online scenario \citep{COMET}. To address this limitation, we proposed a novel approach called Contrastive Mean Teacher (COMET) tailored to the online setting. It uses a mean teacher for pseudo-labeling and applies a combination of a contrastive loss and an entropy loss. Recently, we introduced an alternative pseudo-labeling technique leveraging a Gaussian Mixture Model (GMM), which significantly improves memory-efficiency \citep{GMM}. Additionally, we replaced the entropy loss with a KL-divergence loss.

\section{Method}
\subsection{Preliminaries}
The foundation of DA is a model $f_\mathrm{s}$ pre-trained on the source dataset $\mathcal{D}_\mathrm{s}=\{\boldsymbol{x}_i^\mathrm{s}\in\mathcal{X}_\mathrm{s}, y_i^\mathrm{s}\in\mathcal{Y}_\mathrm{s}\}_{i=1}^{N_\mathrm{s}}$. This source model comprises a feature extractor $g_s$ and a classifier $h_s$, such that $f_s=h_s\circ g_s$. The objective of DA is to adapt $f_s$ to an unlabeled target domain $\mathcal{D}_\mathrm{t}=\{\boldsymbol{x}_i^\mathrm{t}\in\mathcal{X}_\mathrm{t}, ?\in\mathcal{Y}_\mathrm{t}\}_{i=1}^{N_\mathrm{t}}$. Under the strict source-free constraint, adaptation must proceed without access to the original source dataset $\mathcal{D}_\mathrm{s}$ itself or related information like source prototypes. Moreover, the source training cannot be influenced. As a result, only a standard pre-trained model $f_s$ is available for adaptation.

Unlike classical DA, UniDA additionally accounts for potential category shift between source and target data. As there is typically no prior knowledge of the label-set relationship, UniDA must handle all three different kinds of category shift scenarios, namely PDA ($\mathcal{Y}_\mathrm{t}\subset \mathcal{Y}_\mathrm{s}$), ODA ($\mathcal{Y}_\mathrm{s}\subset \mathcal{Y}_\mathrm{t}$), and OPDA ($\mathcal{Y}_\mathrm{s}\cap\mathcal{Y}_\mathrm{t}\neq\emptyset$, $\mathcal{Y}_\mathrm{s}\nsubseteq\mathcal{Y}_\mathrm{t}$, $\mathcal{Y}_\mathrm{s}\nsupseteq\mathcal{Y}_\mathrm{t}$).

In this work, we address SF-UniDA in a continual setting, where the target data is received as a stream of batches of fixed size $N_b>1$, $\{\boldsymbol{x}_{i, k}^\mathrm{t}\}_{i=1}^{N_\mathrm{b}}$, coming from a sequence of different target domains $\{\mathcal{D}_\mathrm{t}^{(1)}, \mathcal{D}_\mathrm{t}^{(2)}, ...\}$. Here, $i$ indexes samples within a batch and $k$ denotes the batch index. Importantly, each batch contains samples from a single domain, and all batches from one domain are processed sequentially before moving to the next domain. Thereby, the number of batches per domain may vary and is unknown in advance, so domain shifts occur without prior notification. Furthermore, the adaptation must be performed in an online manner, meaning that each batch can only be accessed once and demands an immediate prediction.

\subsection{Mean Teacher}
The use of a mean teacher framework \citep{mean_teacher} is well established in continual test-time adaptation (TTA) \citep{rmt, CoTTA, ccotta}, and has also proven effective in our previously proposed approach for online SF-UniDA \citep{COMET}. It uses two copies of the source model called the student model $f=h\circ g$ and teacher model $\tilde{f}=\tilde{h}\circ \tilde{g}$. During adaptation, the student model is updated via standard backpropagation of the adaptation loss, while the teacher model’s weights $\tilde{\boldsymbol{\delta}}_k$ are updated using an exponential moving average (EMA) of the student’s weights $\boldsymbol{\delta}_k$:
\begin{align}
	\tilde{\boldsymbol{\delta}}_k = \alpha_\mathrm{MT}\tilde{\boldsymbol{\delta}}_{k-1}+(1-\alpha_\mathrm{MT})\boldsymbol{\delta}_k~.
\end{align}
Here, $k$ and $k-1$ denote consecutive time steps (batch indices), and $\alpha_\mathrm{MT}$ is the momentum factor that controls the update speed of the teacher model.

By smoothing the noisy parameter updates of the student, the EMA mechanism in the mean teacher framework helps stabilize learning and mitigates error accumulation during adaptation. Furthermore, by incorporating information from previous iterations through weight averaging, it also helps reduce catastrophic forgetting in long-term continual adaptation. Based on these advantages, we build our pseudo-labeling strategy upon this mean teacher framework.

\subsection{GMM-based pseudo-labeling}
Rather than directly using the teacher’s predictions as pseudo-labels like we do in \citep{COMET}, we apply our GMM-based pseudo-labeling method \citep{GMM} within the teacher’s feature space. This allows us to combine the strengths of the GMM --- such as its ability to accumulate knowledge over the course of online adaptation and its support for a broad range of out-of-distribution (OOD) metrics --- with the stability provided by the mean teacher framework, which is particularly important in the targeted long-term continual adaptation scenario.

The key idea of GMM-based pseudo-labeling is to model the distribution of target data in the feature space as a GMM, where each mode corresponds to one of the $|\mathcal{Y}_\mathrm{s}|$ known source classes. This enables the estimation of the likelihood $p(\boldsymbol{x}_{i,k}^\mathrm{t} \mid c; \hat{\boldsymbol{\mu}}_k(c), \hat{\boldsymbol{\Sigma}}_k(c))$ for a target sample $\boldsymbol{x}_{i,k}^\mathrm{t}$ to belong to a given class $c$. These likelihoods are then used to determine pseudo-labels by selecting the class with the highest likelihood.

\subsubsection{GMM update}
The update of the GMM parameters --- namely, the class-wise mean vectors $\hat{\boldsymbol{\mu}}_{k}(c)$ and covariance matrices $\hat{\boldsymbol{\Sigma}}_{k}(c)$ --- for each incoming target batch $\{\boldsymbol{x}_{i, k}^\mathrm{t}\}_{i=1}^{N_\mathrm{b}}$ is inspired by the Expectation-Maximization (EM) algorithm and consists of two steps. In the E-step, we compute two types of weighting factors. First, each sample’s contribution to the GMM parameters of a specific class is weighted by the teacher model’s softmax prediction for the corresponding class. Second, we apply a batch-wise weighting scheme to ensure that each batch contributes to the GMM parameters proportionally to the number of samples it contains per class. Specifically, a batch that includes more confident predictions for a particular class should have a greater influence on that class’s mean and covariance estimates than a batch with fewer such predictions. To this end, we compute class-wise weights $s_{k}(c)$ by summing the teacher model's classifier outputs, which serve as a soft estimate of class representation:
\begin{align}
	s_{k}(c)= \alpha_\mathrm{GMM}\cdot s_{k-1}(c) + \sum_{i=1}^{N_\mathrm{b}} \tilde{f}_c(\boldsymbol{x}_{i,k}^\mathrm{t})~,
	\label{eq:storage}
\end{align}
where $\tilde{f}_c(\cdot)$ denotes the softmax output of the teacher model corresponding to class $c$ and $\alpha_\mathrm{GMM}\in[0,1]$ is an exponential decay factor controlling the influence of previous batches. The weights are initialized with $s_0(c)=0$ $\forall c$.

In the subsequent M-step, we recursively update the GMM parameters as follows:
\begin{align}
	\hat{\boldsymbol{\mu}}_{k}(c) = \frac{\alpha_\mathrm{GMM}\cdot s_{k-1}(c) \cdot \hat{\boldsymbol{\mu}}_{k-1} \left(c\right) + \sum\limits_{i=1}^{N_\mathrm{b}} \tilde{f}_c(\boldsymbol{x}_{i,k}^\mathrm{t}) \cdot r(\tilde{g}(\boldsymbol{x}_{i,k}^\mathrm{t}))}{s_{k}(c)}~,
	\label{eq:gmmmu}
\end{align}
\begin{align}
	\hat{\boldsymbol{\Sigma}}_{k}(c)= \frac{\left(\splitdfrac{\hspace*{-0.1cm}\alpha_\mathrm{GMM}\cdot s_{k-1}(c) \cdot \hat{\boldsymbol{\Sigma}}_{k-1}(c)+ \sum\limits_{i=1}^{N_\mathrm{b}} \tilde{f}_c(\boldsymbol{x}_{i,k}^\mathrm{t})}{\cdot \left(r(\tilde{g}(\boldsymbol{x}_{i,k}^\mathrm{t}))-\hat{\boldsymbol{\mu}}_{k}(c)\right)\left(r(\tilde{g}(\boldsymbol{x}_{i,k}^\mathrm{t}))-\hat{\boldsymbol{\mu}}_{k}(c)\right)^T}\hspace*{-0.1cm}\right)}{s_{k}(c)}~,
	\label{eq:gmmcov}
\end{align}
where $r(\cdot)$ is a dimensionality reduction function implemented as a linear projection layer trained jointly with the student model. As demonstrated in \citep{GMM}, operating the GMM-based pseudo-labeling in a lower-dimensional space rather than directly in the teacher model's feature space improves both memory efficiency and pseudo-label quality. Empirically, we found that setting the reduced feature dimension to $FD_\mathrm{r} = 64$ achieves a good trade-off between memory efficiency and performance.

For each batch, after updating the GMM parameters, we compute the likelihood $p(\boldsymbol{x}_{i,k}^\mathrm{t} \mid c; \hat{\boldsymbol{\mu}}_k(c), \hat{\boldsymbol{\Sigma}}_k(c))$ for each target sample with respect to all known classes $1 \leq c \leq |\mathcal{Y}_\mathrm{s}|$. The final pseudo-label assignment is then determined by selecting the class with the highest likelihood, unless the sample is identified as OOD based on the criteria introduced in the following subsection.

\subsubsection{Out-of-distribution (OOD) detection}
While selecting the class with the highest likelihood from the GMM provides a natural way to assign pseudo-labels, it is essential for UniDA to detect samples that do not belong to any of the known classes and should instead be labeled as unknown. We address this by computing an OOD score $o(\boldsymbol{x}_{i,k}^\mathrm{t})$ for each target sample and applying a threshold-based decision rule for pseudo-labeling. As shown in our previous work \citep{Analysis_Pseudo_Labeling}, the accuracy of pseudo-labels is significantly more critical for successful adaptation than the number of samples used. Accordingly, we apply two thresholds, the lower bound $\tau_\mathrm{l}$ and the upper bound $\tau_\mathrm{u}$, to filter out uncertain samples with intermediate OOD scores, thereby reducing the risk of noisy updates. The resulting pseudo-labeling rule is given by:
\begin{align}
	\hat{y}_{i,k}^\mathrm{pl}=\begin{cases}
		\arg\max\limits_{c\in\mathcal{Y}_\mathrm{s}} {\boldsymbol{p}}_{i,k}
		& o({\boldsymbol{x}}^\mathrm{t}_{i,k})\leq\tau_\mathrm{l}\\
		|\mathcal{Y}_\mathrm{s}|+1 & o({\boldsymbol{x}}^\mathrm{t}_{i,k})\geq\tau_\mathrm{u}\\
		\text{ignored} & \text{otherwise}
	\end{cases}
\end{align}
Here, $\boldsymbol{p}_{i,k} = \left[ p(\boldsymbol{x}_{i,k}^\mathrm{t} | c; \hat{\boldsymbol{\mu}}_{k}(c), \hat{\boldsymbol{\Sigma}}_{k}(c)) \right]_{c=1}^{|\mathcal{Y}_\mathrm{s}|}$ denotes the vector of class-wise likelihoods obtained from the GMM and $\hat{y}_{i,k}^\mathrm{pl}$ is the pseudo-label for sample $\boldsymbol{x}^t_{i,k}$. Moreover, we use $|\mathcal{Y}_\mathrm{s}|+1$ as class index for the unknown class.

Rather than manually setting the OOD score thresholds $\tau_\mathrm{l}$ and $\tau_\mathrm{u}$, we determine them adaptively during the first $N_\mathrm{init}$ target batches. For each of these batches, we compute the OOD score distribution $\{o(\boldsymbol{x}_{i,k}^\mathrm{t})\}_{i=1}^{N_b}$ and extract the boundary scores separating the top and bottom $100\cdot (1 - p_\mathrm{reject})/2$ percent of samples with the highest and lowest OOD scores, respectively. Hence, these boundaries are given by the $(1-(1-p_\mathrm{reject})/2)$-quantile and the $( (1-p_\mathrm{reject})/2)$-quantile of the batch's OOD score values. The resulting per-batch quantiles are then averaged across the first $N_\mathrm{init}$ target batches to obtain the final thresholds $\tau_\mathrm{l}$ and $\tau_\mathrm{u}$, which remain fixed thereafter. This strategy offers intuitive control through the interpretable parameter $p_\mathrm{reject}$, which specifies the fraction of uncertain samples excluded from adaptation, without requiring direct tuning of the raw threshold values.

The GMM-based modeling of the feature space naturally supports a variety of metrics for computing the OOD score $o(\boldsymbol{x}_{i,k}^\mathrm{t})$. In this work, we focus on two complementary metrics that we found to perform best across different datasets:
\begin{itemize}
	\item \textbf{Mahalanobis distance:} This score measures how far a sample’s feature representation lies from the center of its closest class-conditional Gaussian component, effectively capturing class-consistency in the feature space. It is defined as:
	\begin{align}
		o_\mathrm{m}(\boldsymbol{x}_{i,k}^\mathrm{t}) = \min_{c \in \mathcal{Y}_\mathrm{s}} \left( r(\tilde{g}(\boldsymbol{x}_{i,k}^\mathrm{t})) - \hat{\boldsymbol{\mu}}_k(c) \right)^\top \hspace*{-0.1cm} \hat{\boldsymbol{\Sigma}}_k(c)^{-1}\hspace*{-0.1cm} \left( r(\tilde{g}(\boldsymbol{x}_{i,k}^\mathrm{t})) - \hat{\boldsymbol{\mu}}_k(c) \right).
	\end{align}
	
	\item \textbf{Normalized entropy of the likelihoods:} This score measures the uncertainty of a sample's class membership based on the class-wise likelihoods computed from the GMM. High entropy indicates ambiguity in the class assignment and thus a higher likelihood of the sample being out-of-distribution. It is computed as:
	\begin{align}
		o_\mathrm{e}(\boldsymbol{x}_{i,k}^\mathrm{t}) = -\frac{1}{\log |\mathcal{Y}_\mathrm{s}|}\cdot\boldsymbol{p}_{i,k}^T\cdot\log \boldsymbol{p}_{i,k}.
	\end{align}
\end{itemize}

We empirically found that the Mahalanobis distance performs better on datasets with a small number of known classes where the feature clusters are well-separated, while entropy tends to be more effective in scenarios with overlapping class distributions or higher uncertainty. The choice of metric can be adapted depending on the application scenario.

\subsection{Contrastive loss}
Our recent analysis \citep{Analysis_Pseudo_Labeling} demonstrated that contrastive loss is more robust to noisy or imperfect pseudo-labels than cross-entropy loss in the context of UniDA via self-training. Moreover, it naturally complements GMM-based pseudo-labeling by promoting compact and well-separated feature clusters for the known classes. Accordingly, we continue to employ contrastive loss, which has already proven effective in our prior work on online SF-UniDA \citep{COMET, GMM}.

To construct the contrastive loss, we form multiple positive and negative pairs for each target sample in a batch that received a pseudo-label of a known class. Positive pairs are defined in two ways: first, by pairing the sample with all other target samples that share the same pseudo-label, and second, by pairing it with the GMM mean corresponding to that class. Negative pairs are formed with all remaining target samples, including those pseudo-labeled as unknown, as well as with all other GMM class means. Thereby, we extend the data by adding an augmentation of each sample using the augmentations from \citep{CoTTA}. This strategy not only increases the number of training instances but also encourages the model to learn a more stable and transformation-invariant feature space. Overall, this results in the following loss function:

\begin{equation}
	\label{eq:contrastive_loss}
	\begin{aligned}
		\mathcal{L}_{\mathrm{c}}\hspace*{-0.05cm} =\hspace*{-0.05cm} &- \hspace*{-0.04cm} \sum_{j=1}^{2N_\mathrm{b}} \sum_{i=1}^{2N_\mathrm{b}} \mathds{1}(\hat{y}_{j,k}^{\mathrm{pl}}\hspace*{-0.07cm}=\hspace*{-0.05cm}\hat{y}_{i,k}^\mathrm{pl}\wedge\hat{y}_{i,k}^{\mathrm{pl}}\in\mathcal{Y}_\mathrm{s}) \log \frac{\exp\hspace*{-0.07cm}\left(\hspace*{-0.04cm}\frac{\langle r(g(\breve{\boldsymbol{x}}_{j,k}^\mathrm{t})),r(g(\breve{\boldsymbol{x}}_{i,k}^\mathrm{t}))\rangle}{\tau}\hspace*{-0.04cm}\right)}{\sum\limits_{l=1}^{2N_\mathrm{b}}\hspace*{-0.05cm}\exp \hspace*{-0.07cm}\left(\hspace*{-0.04cm}\frac{\langle r(g(\breve{\boldsymbol{x}}_{l,k}^\mathrm{t})),r(g(\breve{\boldsymbol{x}}_{i,k}^\mathrm{t}))\rangle}{\tau}\hspace*{-0.04cm}\right)} \\
		&-\hspace*{-0.04cm} \sum_{c=1}^{|\mathcal{Y}_\mathrm{s}|} \sum_{i=1}^{2N_\mathrm{b}} \mathds{1}(\hat{y}_{i,k}^\mathrm{pl}\hspace*{-0.07cm}=\hspace*{-0.05cm}c) \log \frac{\exp\hspace*{-0.07cm}\left(\hspace*{-0.04cm}\frac{\langle\hat{\boldsymbol{\mu}}_{k}\hspace*{-0.05cm}(c),r(g(\breve{\boldsymbol{x}}_{i,k}^\mathrm{t}))\rangle}{\tau}\hspace*{-0.04cm}\right)}{\sum\limits_{c=1}^{|\mathcal{Y}_\mathrm{s}|}\hspace*{-0.05cm}\exp\hspace*{-0.07cm} \left(\hspace*{-0.04cm}\frac{\langle\hat{\boldsymbol{\mu}}_{k}\hspace*{-0.05cm}(c),r(g(\breve{\boldsymbol{x}}_{i,k}^\mathrm{t}))\rangle}{\tau}\hspace*{-0.04cm}\right)}~.
	\end{aligned}
\end{equation}
Here, $\{\breve{\boldsymbol{x}}_{i,k}^\mathrm{t}\}_{i=1}^{2N_\mathrm{b}}=\{\boldsymbol{x}_{i,k}^\mathrm{t}\}_{i=1}^{N_\mathrm{b}}\cup\{\bar{\boldsymbol{x}}_{i,k}^\mathrm{t}\}_{i=1}^{N_\mathrm{b}}$ is the union of the original target samples $\{\boldsymbol{x}_{i, k}^\mathrm{t}\}_{i=1}^{N_\mathrm{b}}$ and their augmented counterparts $\{\bar{\boldsymbol{x}}_{i,k}^\mathrm{t}\}_{i=1}^{N_\mathrm{b}}$. Moreover, $\mathds{1}(\cdot)$ is the indicator function, $\langle\cdot\rangle$ denotes cosine similarity, and $\tau$ is the temperature parameter.

\subsection{Entropy loss}
To complement the contrastive loss, which exclusively updates the feature extractor, we apply an additional loss to also refine the classifier's ability to translate features into accurate predictions. This is particularly important for identifying unknown samples with an entropy-based OOD score, as the classifier was originally trained in a closed-set setting and may not reliably produce high-entropy outputs for out-of-distribution inputs. To address this, we adopt the entropy-based loss from our previous work \citep{COMET}, encouraging confident predictions (low entropy) for samples pseudo-labeled as known and uncertain predictions (high entropy) for those labeled as unknown:
\begin{align}
	\mathcal{L}_\mathrm{e}=\frac{1}{N_b}\sum_{i=1}^{N_b}\mathds{1}(\hat{y}_{i,k}^{\mathrm{pl}}\in\mathcal{Y}_s)\cdot I(\boldsymbol{x}_{i,k}^\mathrm{t}) - \frac{1}{N_b}\sum_{i=1}^{N_b} \mathds{1}(\hat{y}_{i,k}^{\mathrm{pl}}=|\mathcal{Y}_s|+1)\cdot I(\boldsymbol{x}_{i,k}^\mathrm{t})~,
\end{align}
where the normalized entropy is computed from the classwise softmax outputs of the student model $f_c(\cdot)$ to learn its models weights $\boldsymbol{\delta}_k$:
\begin{align}
	I(\boldsymbol{x}_{i,k}^\mathrm{t})=-\frac{1}{\log |\mathcal{Y}_\mathrm{s}|}\cdot\sum_{c=1}^{|\mathcal{Y}_\mathrm{s}|} f_c(\boldsymbol{x}_{i,k}^\mathrm{t})\cdot\log f_c(\boldsymbol{x}_{i,k}^\mathrm{t})~.
	\label{eq:entropy}
\end{align}

\subsection{Consistency losses}
A major challenge in continual SF-UniDA arises from the extended adaptation timeline, which can lead the model to drift too far from the source model. While some drift is necessary to adapt to the target domain, excessive drift can destabilize training and increases the risk of overfitting to the incoming target data, potentially causing catastrophic forgetting or model collapse. This risk is particularly pronounced when one of the target domains features a large domain gap leading to a low pseudo-label quality. In such cases, due to the reinforcement of wrong predictions, the model may not be able to recover even if the subsequent domain shift is less severe. To address this and ensure stable and effective adaptation, we introduce consistency losses that constrain the deviation from the source model and promote coherence between student and teacher.

\paragraph{Source Consistency Loss}  
To limit divergence from the original source model, we enforce consistency between the source feature extractor \(g_s\) and the student feature extractor \(g\):
\begin{align}
	\mathcal{L}_\mathrm{con,src} = \frac{1}{N_\mathrm{b}}\sum_{i=1}^{N_\mathrm{b}}\left\| g(\boldsymbol{x}_{i,k}^\mathrm{t}) - g_s(\boldsymbol{x}_{i,k}^\mathrm{t}) \right\|_2.
\end{align}

\paragraph{Student–Teacher Consistency Loss}  
In addition, to stabilize the mean teacher framework, we enforce consistency between the feature extractor outputs of student and teacher, preventing the models from drifting too far apart:
\begin{align}
	\mathcal{L}_\mathrm{con,mt} = \frac{1}{N_\mathrm{b}}\sum_{i=1}^{N_\mathrm{b}}\left\| g(\boldsymbol{x}_{i,k}^\mathrm{t}) - \tilde{g}(\boldsymbol{x}_{i,k}^\mathrm{t}) \right\|_2.
\end{align}

\paragraph{Overall Loss Function.}  
The final training objective of GMM-COMET integrates both consistency losses with the contrastive loss \(\mathcal{L}_\mathrm{c}\) and the entropy loss \(\mathcal{L}_\mathrm{ent}\), leading to the total loss:
\begin{align}
	\mathcal{L} = \mathcal{L}_\mathrm{c} + \lambda_1 \mathcal{L}_\mathrm{e} + \lambda_2 \mathcal{L}_\mathrm{con,src} + \lambda_3 \mathcal{L}_\mathrm{con,mt}.
\end{align}

\subsection{Inference}
For inference, we apply the same OOD detection strategy used for pseudo-labeling, but with a single unified threshold $\tau$ computed as the average of the two thresholds used during pseudo-labeling: $\tau=(\tau_\mathrm{l}+\tau_\mathrm{u})/2$. For the prediction of known classes, we ensemble the outputs of the student and teacher models to increase robustness. The final prediction is given by:
\begin{align}
	\label{eq:inference}
	\hat{y}_{i,k} = \begin{cases}
		\arg\max\limits_c f_c(\boldsymbol{x}_{i,k}^\mathrm{t})+\tilde{f}_c(\boldsymbol{x}_{i,k}^\mathrm{t}) & o(\boldsymbol{x}_{i,k}^\mathrm{t})\leq \tau\\
		|\mathcal{Y}_\mathrm{s}|+1 & o(\boldsymbol{x}_{i,k}^\mathrm{t})> \tau\end{cases}~.
\end{align}

\section{Experiments}
\subsection{Setup}
\subsubsection{Datasets}
We evaluate our method on the public DA datasets DomainNet \citep{domainnet}, 
CIFAR-10-C and CIFAR-100-C \citep{corrupted_datasets}. From DomainNet, we utilize the domains clipart (c), painting (p), real (r), and sketch (s), each comprising approximately 48,000 to 173,000 images, distributed across 345 classes. 
CIFAR-10-C and CIFAR-100-C are corruption-augmented extensions of the original CIFAR-10 and CIFAR-100 datasets \citep{CIFAR}, with their respective training splits of 60,000 images serving as the source domains. These corruption-augmented datasets retain the same 10 and 100 classes across 10,000 testing images as the originals. Each of the 15 common corruption types (e.g., noise, blur, and weather effects) is applied to all 10,000 images at five different severity levels, resulting in 150,000 corrupted images per dataset per severity level. In this work, we focus exclusively on the most challenging severity level 5.

To construct the continual shift scenario, we arrange the target domains in a fixed, predefined order. For DomainNet, we use all domains except the chosen source domain as target domains and sort them alphabetically. For CIFAR-10-C and CIFAR-100-C, we follow the original corruption order introduced in \citep{corrupted_datasets}, consistent with prior work on continual TTA like \cite{CoTTA, rmt}.

Before concatenation, each domain is internally shuffled to remove any sample ordering effects, but the domains themselves remain contiguous. During adaptation, batches are drawn sequentially from this concatenated dataset without further shuffling. Consequently, each batch contains samples from exactly one domain, and the model processes all samples of one target domain before moving on to the subsequent domain, thereby simulating a realistic continual domain shift.

In this work, we assume that all consecutive target domains share the same label space, i.e., $\mathcal{Y}_t=\mathcal{Y}_t^{(1)}=\mathcal{Y}_t^{(2)}=\cdots$. We leave the analysis of changing category shifts in addition to (or instead of) the continual domain shifts open for future work. The class splits we used to construct the category shifts for the three datasets are shown in Table \ref{tab:class_splits}. We take the first $|\mathcal{Y}_\mathrm{s}|$ classes as source classes and the last $|\mathcal{Y}_\mathrm{t}|$ classes as target classes. The overlapping $|\mathcal{Y}_\mathrm{s}\cap\mathcal{Y}_\mathrm{t}|$ classes are shared between the source and target domains. Thereby, the ordering of the classes is alphabetically for both DomainNet and CIFAR-10-C while CIFAR-100-C follows an ordering according to superclasses which can be found in \citep{krizhevsky2009learning}. 

\begin{table}
	\caption{Class splits, i.e. number of the shared, source-private and target-private classes, for OPDA, ODA and PDA, respectively}
	\label{tab:class_splits}
	\begin{tabularx}{\linewidth}{l*{3}{c}}
		\toprule
		& \multicolumn{3}{c}{$|\mathcal{Y}_\mathrm{s}\cap\mathcal{Y}_\mathrm{t}|$, $|\mathcal{Y}_\mathrm{s}\backslash\mathcal{Y}_\mathrm{t}|$, $|\mathcal{Y}_\mathrm{t}\backslash\mathcal{Y}_\mathrm{s}|$}\\
		\cmidrule(lr){2-4}
		& PDA & ODA & OPDA\\
		\midrule
		DomainNet & 200, 145, 0 & 200, 0, 145 & 150, 50, 145 \\
		CIFAR-10-C & 60, 40, 0 & 60, 0, 40 & 40, 20, 40 \\
		CIFAR-100-C & 6, 4, 0 & 6, 0, 4 & 4, 2, 4 \\
		\bottomrule
	\end{tabularx}
\end{table}

\subsubsection{Competing methods}
To ensure a fair comparison, all competing methods are evaluated under the same continual, source-free, and universal domain adaptation setting. As a baseline, we include the source model without any adaptation, using an entropy threshold to reject unknown samples. This baseline establishes a lower bound that adaptation methods must surpass to demonstrate their effectiveness. Next, we evaluate several established approaches originally developed for offline SF-UniDA, including GLC \citep{glc}, GLC++ \citep{glc++}, and LEAD \citep{lead}, which we adapt to the continual setting by applying them batch-wise. We further include the SHOT-O variant of SHOT \citep{shot}, originally proposed for ODA. Although not designed for universal DA, we apply SHOT-O across all three scenarios (PDA, ODA, and OPDA) to ensure comparability. Additionally, we evaluate OWTTT \citep{owttt}, a method developed for robust OOD detection in online TTA, which we repurpose for continual SF-UniDA. Finally, we include our two recently proposed approaches specifically targeting online SF-UniDA — the COMET variants COMET-P and COMET-F \citep{COMET}, as well as the GMM-based method \citep{GMM} — all of which are directly applicable in the continual setting without modification. 

Note that both OWTTT and COMET-P require access to source prototypes, which, depending on the definition, may violate the source-free assumption and limits their applicability in practical settings.

\subsubsection{Implementation}
To ensure consistent and meaningful comparisons, all methods are evaluated using the same pre-trained source model. We adopt a ResNet-50 \citep{he2016deep} backbone initialized with ImageNet weights \citep{deng2009imagenet}, followed by a $256$-dimensional feature projection head and a classification layer. The source training is carried out the same way like originally introduced by \citep{shot} and adopted by \citep{glc, glc++, lead, COMET, GMM}. For adaptation, we use SGD with momentum of $0.9$, a learning rate of $0.001$ and a batch size of $N_b=64$ for all datasets. Both the momentum factors for updating the EMA-based teacher and the GMM parameters are set to $\alpha_\mathrm{MT}=\alpha_\mathrm{GMM}=0.999$ for DomainNet and $\alpha_\mathrm{MT}=\alpha_\mathrm{GMM}=0.99$ for the two CIFAR datasets. To apply the GMM, the original features are projected from $\mathbb{R}^{256}$ onto a subspace with $FD_\mathrm{r} = 64$ dimensions. In terms of OOD detection, we apply to the Mahalanibis distance as OOD score for CIFAR-10 and the normalized entropy of the GMM-based likelihoods for DomainNet and CIFAR-100. For initializing the thresholds used in OOD detection, we collect statistics over the first $N_\mathrm{init} = 50$ target batches and set the rejection ratio to $p_\mathrm{reject} = 0.5$ for CIFAR-10 and CIFAR-100 and to $p_\mathrm{reject} = 0.65$ for DomainNet. Regarding the contrastive loss, we apply a temperature of $\tau=0.1$. The overall loss function is composed of three components weighted by factors $\lambda_1$, $\lambda_2$, and $\lambda_3$. We fix $\lambda_1 = 1$ for all datasets. For DomainNet and CIFAR-100, we choose $\lambda_2 = 2$ and $\lambda_3 = 1$, while for CIFAR-10 we set $\lambda_2 = 5$ and $\lambda_3 = 2$ to stay consistent with the strong source-only baseline and ensure stable adaptation.

\subsubsection{Metrics}
As customary, we use the H-score as the evaluation metric for ODA and OPDA. It is defined as the harmonic mean of the known-class accuracy $\mathrm{acc}_k$ and the unknown-class accuracy $\mathrm{acc}_u$:
\begin{align}
	\text{H-score} = \frac{2 \cdot acc_k \cdot acc_u}{acc_k + acc_u}\,.
\end{align}
Here, $\mathrm{acc}_k$ and $\mathrm{acc}_u$ denote the accuracies computed only on samples whose ground-truth labels belong to the known classes and to the unknown class, respectively.

Since the PDA scenario contains only known classes, we use the (known-class) accuracy as the evaluation metric in this scenario instead.

For all scenarios, we first compute the metric separately on each consecutive target domain and then report the average across domains. This ensures that all target domains, and thus all domain shifts, are weighted equally, rather than proportionally to their number of batches.

\begin{table*}
	\caption{Accuracies in $\%$ for PDA. Best results are in highlighted in bold while results worse compared to the source-only performance are marked in red. D(c), D(p), D(r), and D(s) denote DomainNet with clipart, painting, real, and sketch as the source domains, respectively. C10 and C100 refer to CIFAR-10-C and CIFAR-100-C.}
	\label{tab:results_PDA}
	\begin{tabularx}{\textwidth}{p{0.8cm} *{10}{c} }
		\toprule
		PDA & \parbox[t]{4mm}{\rotatebox[origin=c]{90}{Source-only}} & \parbox[t]{4mm}{\rotatebox[origin=c]{90}{OWTTT}} & \parbox[t]{4mm}{\rotatebox[origin=c]{90}{COMET-P}} & \parbox[t]{4mm}{\rotatebox[origin=c]{90}{SHOT-O}} & \parbox[t]{4mm}{\rotatebox[origin=c]{90}{GLC}}& \parbox[t]{4mm}{\rotatebox[origin=c]{90}{GLC++}} & \parbox[t]{4mm}{\rotatebox[origin=c]{90}{LEAD}} & \parbox[t]{4mm}{\rotatebox[origin=c]{90}{COMET-F}} & \parbox[t]{4mm}{\rotatebox[origin=c]{90}{GMM}} & \parbox[t]{4mm}{\rotatebox[origin=c]{90}{Ours}}\\
		\midrule
		D(c) & $23.3$ & $\color{red}22.4$ & $32.9$ & $29.9$ & $\color{red}7.4$ & $\color{red}8.3$ & $\color{red}2.4$ & $31.3$ & $34.6$ & \boldmath$35.2$\\
		D(p) & $30.1$ & $\color{red}25.5$ & \boldmath$40.5$ & $38.0$ & $\color{red}9.4$ & $\color{red}10.1$ & $\color{red}5.2$ & $38.4$ & $38.6$ & $39.1$\\
		D(r) & $29.1$ & $\color{red}21.6$ & $34.3$ & \boldmath$40.0$ & $\color{red}7.7$ & $\color{red}8.0$ & $\color{red}3.1$ & $34.2$ & $37.2$ & $37.7$\\
		D(s) & $22.6$ & $\color{red}15.5$ & $37.6$ & $33.4$ & $\color{red}12.3$ & $\color{red}12.8$ & $\color{red}6.6$ & $35.0$ & $37.7$ & \boldmath$39.7$\\
		C100 & $19.1$ & $22.8$ & $23.8$ & $21.9$ & $25.1$ & $27.3$ & $\color{red}18.7$ & $22.8$ & $31.8$ & \boldmath$38.4$\\
		C10 & $46.2$ & $\color{red}39.8$ & $51.2$ & $48.9$ & $56.1$ & $\color{red}42.4$ & $51.2$ & $47.4$ & $\color{red}36.3$ & \boldmath$59.9$\\
		
		\bottomrule
	\end{tabularx}
\end{table*}
\vspace*{0mm}
\begin{table*}
	\caption{H-scores in $\%$ for ODA. Best results are in highlighted in bold while results worse compared to the source-only performance are marked in red. D(c), D(p), D(r), and D(s) denote DomainNet with clipart, painting, real, and sketch as the source domains, respectively. C10 and C100 refer to CIFAR-10-C and CIFAR-100-C.}
	\label{tab:results_ODA}
	\begin{tabularx}{\textwidth}{p{0.8cm} *{10}{c} }
		\toprule
		ODA & \parbox[t]{4mm}{\rotatebox[origin=c]{90}{Source-only}} & \parbox[t]{4mm}{\rotatebox[origin=c]{90}{OWTTT}} & \parbox[t]{4mm}{\rotatebox[origin=c]{90}{COMET-P}} & \parbox[t]{4mm}{\rotatebox[origin=c]{90}{SHOT-O}} & \parbox[t]{4mm}{\rotatebox[origin=c]{90}{GLC}}& \parbox[t]{4mm}{\rotatebox[origin=c]{90}{GLC++}} & \parbox[t]{4mm}{\rotatebox[origin=c]{90}{LEAD}} & \parbox[t]{4mm}{\rotatebox[origin=c]{90}{COMET-F}} & \parbox[t]{4mm}{\rotatebox[origin=c]{90}{GMM}} & \parbox[t]{4mm}{\rotatebox[origin=c]{90}{Ours}}\\
		\midrule
		D(c) & $42.5$ & $\color{red}39.3$ & $47.7$ & $46.8$ & $\color{red}20.0$ & $\color{red}20.0$ & $\color{red}27.3$ & $46.0$ & $47.9$ & \boldmath$48.3$\\
		D(p) & $46.5$ & $\color{red}42.1$ & $50.1$ & \boldmath$50.3$ & $\color{red}17.5$ & $\color{red}17.8$ & $\color{red}26.9$ & $48.9$ & $48.8$ & $48.9$\\
		D(r) & $45.2$ & $\color{red}36.7$ & \boldmath$50.0$ & $48.9$ & $\color{red}15.5$ & $\color{red}15.0$ & $\color{red}23.6$ & $49.5$ & $47.9$ & $47.4$\\
		D(s) & $40.3$ & $\color{red}38.4$ & $49.2$ & $45.5$ & $\color{red}25.4$ & $\color{red}24.9$ & $\color{red}32.8$ & $47.2$ & $50.0$ & \boldmath$51.2$\\
		C100 & $35.1$ & $41.8$ & $38.5$ & $36.9$ & $44.1$ & $44.4$ & $38.0$ & $37.7$ & $43.8$ & \boldmath$44.3$\\
		C10 & $46.5$ & $\color{red}44.5$ & $\color{red}45.0$ & $\color{red}46.3$ & $\color{red}42.7$ & $\color{red}24.1$ & $\color{red}43.8$ & $\color{red}44.6$ & $50.3$ & \boldmath$50.6$\\
		
		\bottomrule
	\end{tabularx}
\end{table*}

\begin{table*}
	\caption{H-scores in $\%$ for OPDA. Best results are in highlighted in bold while results worse compared to the source-only performance are marked in red. D(c), D(p), D(r), and D(s) denote DomainNet with clipart, painting, real, and sketch as the source domains, respectively. C10 and C100 refer to CIFAR-10-C and CIFAR-100-C.}
	\label{tab:results_OPDA}
	\begin{tabularx}{\textwidth}{p{0.8cm} *{10}{c} }
		\toprule
		OPDA & \parbox[t]{4mm}{\rotatebox[origin=c]{90}{Source-only}} & \parbox[t]{4mm}{\rotatebox[origin=c]{90}{OWTTT}} & \parbox[t]{4mm}{\rotatebox[origin=c]{90}{COMET-P}} & \parbox[t]{4mm}{\rotatebox[origin=c]{90}{SHOT-O}} & \parbox[t]{4mm}{\rotatebox[origin=c]{90}{GLC}}& \parbox[t]{4mm}{\rotatebox[origin=c]{90}{GLC++}} & \parbox[t]{4mm}{\rotatebox[origin=c]{90}{LEAD}} & \parbox[t]{4mm}{\rotatebox[origin=c]{90}{COMET-F}} & \parbox[t]{4mm}{\rotatebox[origin=c]{90}{GMM}} & \parbox[t]{4mm}{\rotatebox[origin=c]{90}{Ours}}\\
		\midrule
		D(c) & $45.9$ & $\color{red}44.3$ & $50.7$ & $50.0$ & $\color{red}25.0$ & $\color{red}24.9$ & $\color{red}28.9$ & $49.3$ & $50.8$ & \boldmath$51.1$\\
		D(p) & $46.1$ & $\color{red}39.7$ & \boldmath$50.9$ & $49.9$ & $\color{red}21.9$ & $\color{red}21.6$ & $\color{red}28.1$ & $49.6$ & $50.1$ & $49.9$\\
		D(r) & $45.9$ & $\color{red}38.6$ & \boldmath$51.0$ & $49.4$ & $\color{red}21.3$ & $\color{red}21.0$ & $\color{red}24.6$ & $50.9$ & $49.5$ & $48.8$\\
		D(s) & $42.3$ & $\color{red}31.8$ & $50.2$ & $46.8$ & $\color{red}30.3$ & $\color{red}30.2$ & $\color{red}33.3$ & $48.5$ & $51.4$ & \boldmath$52.5$\\
		C100 & $35.0$ & $40.3$ & $38.6$ & $36.9$ & $43.2$ & $44.0$ & $36.3$ & $37.4$ & $43.8$ & \boldmath$44.3$\\
		C10 & $46.8$ & $\color{red}39.5$ & $51.5$ & $47.7$ & $\color{red}38.4$ & $\color{red}15.1$ & $47.8$ & $51.5$ & \boldmath$52.4$ & $47.2$\\
		\bottomrule
	\end{tabularx}
\end{table*}

\subsection{Results}
Table \ref{tab:results_PDA}, Table \ref{tab:results_ODA}, and Table \ref{tab:results_OPDA} summarize the results. Similar to the online scenario \citep{COMET, GMM}, GLC, GLC++, and LEAD perform poorly in the continual setting due to their reliance on clustering and kNN. Interestingly, OWTTT underperforms the source-only baseline in almost all scenarios, despite exhibiting competitive performance in the online setting. This degradation is likely caused by outdated prototypes, particularly for unknown classes, which become misaligned after each additional domain shift.

While COMET-P, COMET-F, SHOT, and GMM achieve competitive results in some domains, their performance varies notably across tasks. In contrast, our method GMM-COMET delivers consistently strong results across all domain shifts, remaining close to or achieving the best performance in nearly all settings. It attains the highest results for all three category shifts on CIFAR-100 and for DomainNet with sketch and clipart as source domains. For CIFAR-10, GMM-COMET is the only method that consistently improves upon the source-only baseline across all category shifts, clearly outperforming competing approaches in both ODA and PDA scenarios. Although it is slightly outperformed in DomainNet with painting and real as source domains, it still demonstrates highly competitive results. Overall, GMM-COMET proves to be the most reliable and consistent approach, combining strong adaptation capability with robustness across diverse domain shifts.

\subsection{Ablation studies}
To isolate the effects of the newly introduced components compared to our GMM-based baseline method \citep{GMM}, we focus the ablation analysis on the two main modifications, namely the consistency losses and the mean teacher. All other elements remain identical to the original method. We perform these ablation experiments on CIFAR-100-C and DomainNet with sketch as source domain under the PDA and OPDA scenarios, respectively.

\begin{figure}[!t]
	\centering
	\includegraphics[width=0.9\linewidth]{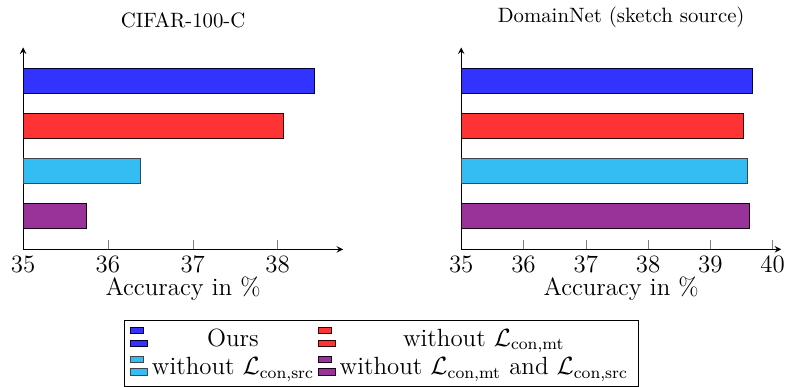}
	\caption{Impact of the consistency losses in the PDA scenario.}
	\label{fig:reg_loss_PDA}
\end{figure}
\begin{figure}[!t]
	\centering
	\includegraphics[width=0.9\linewidth]{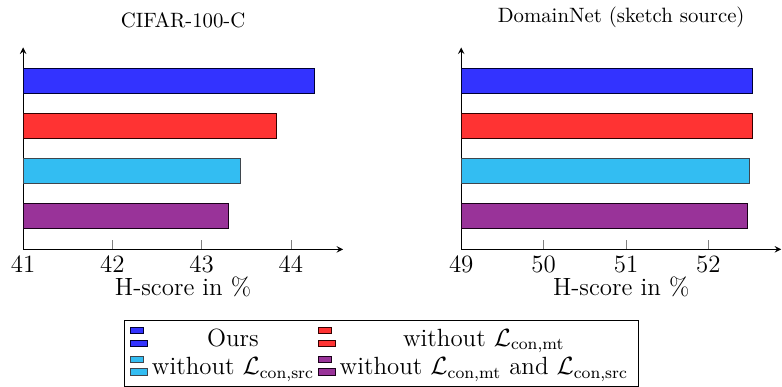}
	\caption{Impact of the consistency losses in the OPDA scenario.}
	\label{fig:reg_loss_OPDA}
\end{figure}
 
\subsubsection{Impact of consistency losses}
Figure \ref{fig:reg_loss_PDA} and Figure \ref{fig:reg_loss_OPDA} report the impact of removing the consistency losses. On CIFAR-100-C, both the student–teacher consistency loss $\mathcal{L}_\mathrm{con,mt}$ and the source consistency loss $\mathcal{L}_\mathrm{con,src}$ provide clear benefits under both PDA and OPDA. The effect is particularly pronounced in the PDA setting, where jointly applying both losses yields an improvement of nearly $3\%$. In contrast, on DomainNet the influence of the consistency losses is noticeably weaker. Across both PDA and OPDA, their removal results in no or only marginal performance changes, indicating that the model is less sensitive to additional regularization in this setting.
Overall, these results suggest that consistency regularization is especially beneficial in highly dynamic continual shift scenarios such as CIFAR-100-C, which comprises many short, consecutive target domains and thus requires stronger stabilization. In less dynamic settings like DomainNet, characterized by fewer but larger target domains, consistency losses neither significantly improve nor degrade performance. Consequently, their inclusion represents a safe and generally effective strategy for continual SF-UniDA.

\subsubsection{Impact of mean teacher}
\begin{figure}[!t]
	\centering
	\includegraphics[width=0.9\linewidth]{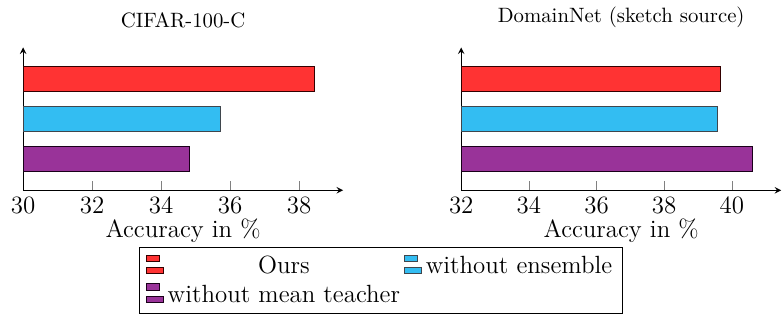}
	\caption{Impact of the mean teacher and corresponding ensembling in the PDA scenario.}
	\label{fig:mt_PDA}
\end{figure}

\begin{figure}[!t]
	\centering
	\includegraphics[width=0.9\linewidth]{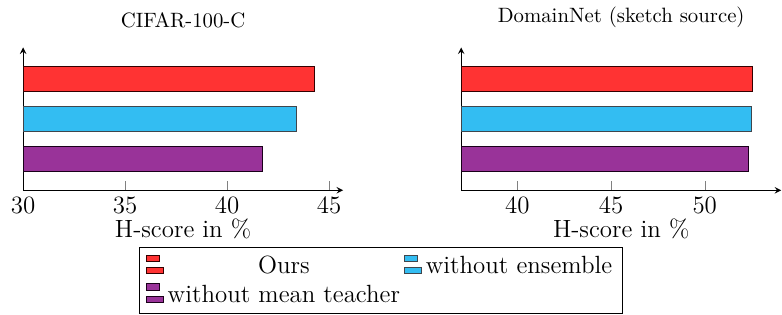}
	\caption{Impact of the mean teacher and corresponding ensembling in the OPDA scenario.}
	\label{fig:mt_OPDA}
\end{figure}

The influence of the mean teacher and ensembling, meaning using the average of the student and teacher outputs for inference, is shown in Figure \ref{fig:mt_PDA} and Figure \ref{fig:mt_OPDA}. Ensembling consistently improves the performance across all scenarios, whereby the benefit is again more pronounced for the more dynamic CIFAR-100-C dataset in contrast to the DomainNet dataset. This indicates that ensembling, similar to the consistency losses, is a safe and effective strategy for continual SF-UniDA.


Completely removing the mean teacher leads to a further significant performance drop on CIFAR-100-C, confirming its stabilizing effect on the pseudo-labeling process under frequent shifts. In contrast, for DomainNet, eliminating the mean teacher causes only a marginal additional performance degradation in the OPDA setting and even yields improved results for PDA compared to the ensembled variant. This suggests that the stabilizing effect of the mean teacher is primarily required in scenarios with dynamic or substantial domain and/or category shifts. In simpler shift settings such as DomainNet PDA, however, this stabilization is unnecessary and the exponential moving average inherent to the mean teacher can instead slow down adaptation. As a result, the mean teacher is not universally beneficial but instead exhibits scenario-dependent effectiveness, proving most useful in the presence of dynamic or challenging shifts.



\section{Conclusion}
In this work, we presented the first systematic study of continual source-free universal domain adaptation (SF-UniDA). Building on our previous approaches towards online SF-UniDA, which served as strong baselines, we adapted and combined their core ideas to explicitly address the challenges of continual domain shifts. Specifically, we integrated GMM-based pseudo-labeling into a mean teacher framework to stabilize adaptation over long target streams and introduced additional consistency losses to further improve robustness. The resulting method, GMM-COMET, is the first approach explicitly tailored to continual SF-UniDA. Through extensive experiments, we provided the first benchmark results for continual SF-UniDA, establishing a reference point for future work in this setting. Across all evaluated scenarios, GMM-COMET is the only method that consistently improves upon the source-only model and, as a result, achieves the best overall performance among all competing approaches. Future work may extend this setting to continual category shifts, either in addition to or instead of continual domain shifts. Another promising direction is to move beyond rejection and enable the classification or clustering of samples from unknown classes, thereby further advancing the scope of SF-UniDA.


\appendix
\label{app1}

  \bibliographystyle{elsarticle-harv} 
  \bibliography{refs}



\end{document}